\documentclass[letterpaper]{article}
\ifdefined\anonymoussubmission
\usepackage[submission]{aaai2027}
\else
\usepackage[preprint]{aaai2027}
\fi
\usepackage[hyphens]{url}
\usepackage{graphicx}
\usepackage{natbib}
\usepackage{caption}
\usepackage{booktabs}
\usepackage{amsmath,amssymb,amsfonts}
\usepackage{algorithm}
\usepackage{algorithmic}
\usepackage{xcolor}
\usepackage{colortbl}
\usepackage{tabularx}

\definecolor{coreRow}{HTML}{EAF2FF}
\definecolor{deltaText}{HTML}{5F6673}
\newcommand{\method}{CoRT}
\newcommand{\gain}[1]{\hspace{0.08em}\textcolor{deltaText}{#1}}
\newcommand{\stackgain}[2]{\shortstack{#1\\\textcolor{deltaText}{#2}}}
\newcommand{\E}{\mathbb{E}}
\newcommand{\sg}{\operatorname{sg}}
\newcolumntype{Y}{>{\centering\arraybackslash}X}

\title{\method: Counterfactual Replay for Token-Level Rubric-Guided Policy Optimization}
\author{
Bo-Wen Zhang\textsuperscript{\rm 1,2,3}\equalcontrib\thanks{Work done during internship at ByteDance.},
Junwei He\textsuperscript{\rm 3}\equalcontrib,
Wen Wang\textsuperscript{\rm 4},
Song-Lin Lv\textsuperscript{\rm 1,2},
Wentao Ma\textsuperscript{\rm 3},
Rongyi Lin\textsuperscript{\rm 3}\thanks{Project lead.},
Shuhan Zhong\textsuperscript{\rm 3},
Lan-Zhe Guo\textsuperscript{\rm 1,2}\corresponding
}
\affiliations{
\textsuperscript{\rm 1}State Key Laboratory of Novel Software Technology, Nanjing University\\
\textsuperscript{\rm 2}School of Intelligence Science and Technology, Nanjing University\\
\textsuperscript{\rm 3}ByteDance \quad
\textsuperscript{\rm 4}University of Chinese Academy of Sciences\\
\url{zbw@smail.nju.edu.cn}, \url{hejunwei@bytedance.com}, \url{guolz@nju.edu.cn}
}

\begin{document}
\maketitle

\begin{abstract}
Rubric-based reinforcement learning enriches language model training by evaluating model outputs against explicit criteria. Yet in GRPO-style pipelines, these structured judgments are reduced to a scalar response-level reward and converted into a response-level advantage, which is broadcast uniformly to all generated tokens. This leaves no explicit mechanism for allocating credit within a response, even when different criteria are grounded in different spans, formatting decisions, or semantic choices.
We propose \method{}, a token-level credit weighting method for rubric-conditioned GRPO. Instead of training an auxiliary token scoring model, \method{} uses counterfactual replay to rescore the same sampled response under the original rubric-conditioned prompt and a matched criteria-free prompt. The resulting tokenwise log-likelihood contrasts serve as a proxy for dependence on the rubric context. \method{} maps these contrasts to bounded, response-normalized weights and uses them to redistribute the signed GRPO advantage across tokens, without introducing an auxiliary scorer or changing the response-level reward.
Experiments across instruction-tuned models and reward granularities show that \method{} improves over matched response-level GRPO in the vast majority of comparisons, with an average gain of 4.4 percentage points. The method remains competitive with learned token-level credit baselines while avoiding a separate relevance-learning stage. These results suggest that policy-internal counterfactual likelihood contrasts provide an effective training signal for within-response credit allocation while retaining the simplicity and stability of GRPO.
\end{abstract}

\section{Introduction}

Reinforcement learning (RL) has become a central component of large language model post-training, enabling models to optimize behaviors that are difficult to capture with supervised fine-tuning alone~\citep{ouyang2022training,jaech2024openai,guo2025deepseek}. This is especially important for open-domain instruction following, where user requests often combine semantic goals with diverse surface and behavioral constraints, and many tasks do not have a single verifiable answer. In many recent RL pipelines, a verifier or reward model assigns a scalar score to each sampled response, and policy optimization methods such as GRPO~\citep{shao2024deepseekmath} update the model using response-level advantages. This formulation is simple and scalable, but it treats a generated response as a single unit of credit assignment, even when different tokens may contribute to different requirements.

Rubric-based RL provides a more structured source of supervision for these settings. Rather than judging a response only by an overall score, a rubric decomposes response behavior into explicit criteria, such as correctness, formatting, or safety~\citep{gunjal2025rubrics,li2026rubrichub}. However, in GRPO-style training, these structured judgments are aggregated into a single advantage before optimization and applied uniformly to all generated tokens. The supervision is structured at the criterion level, while the policy update remains uniform at the token level, leaving potential attribution information encoded in the rubric underused.
A natural way to address this mismatch is to move from response-level reward assignment to token-level credit assignment. In principle, token-level signals can differentiate tokens critical to fulfilling rubric criteria from incidental filler tokens within the final response. One route is to learn an auxiliary token-level relevance or scoring model, but this introduces a separate training stage and an additional component that must be integrated into the RL pipeline. This raises a simpler question: \emph{Can token-level credit be derived directly from policy-internal signals, without training a separate relevance estimator?}

\begin{figure}[t]
\centering
\includegraphics[width=\columnwidth]{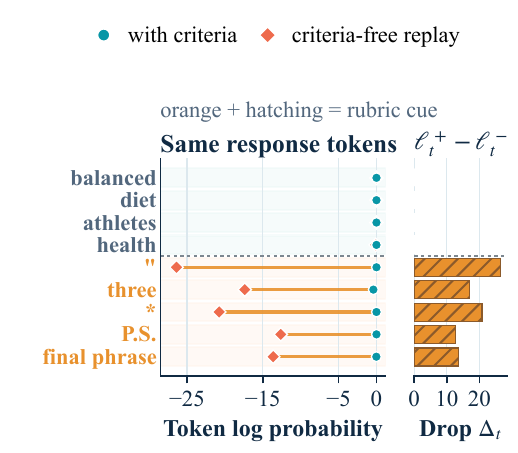}
\caption{Real replay example. The same response is rescored with criteria and under a criteria-free replay. Generic content tokens show little log-probability change, while rubric-controlled tokens (orange) drop more after criteria removal.}
\label{fig:intro-replay-observation}
\end{figure}

Fixed-response replay exposes such a signal. If the same sampled response is rescored after removing the rubric criteria, tokens whose likelihood depends on these criteria tend to change more than generic content tokens. Figure~\ref{fig:intro-replay-observation} illustrates this effect on a real example. Because the response tokens are held fixed, the contrast isolates how the policy's likelihood for the observed trajectory changes under the criteria-free prompt. The highlighted tokens are not labeled by an external annotator; they are exposed by the policy's own response to this local counterfactual prompt change.

Motivated by this observation, we propose \method{} (\textbf{Co}unterfactual \textbf{R}eplay for \textbf{T}oken-level credit weighting), a token-level credit weighting method for rubric-conditioned GRPO. Here, counterfactual replay denotes a fixed-response intervention inspired by replay and counterfactual data augmentation in RL~\citep{andrychowicz2017hindsight,pitis2020counterfactual}: sampled tokens are held fixed and rescored after removing rubric criteria, yielding policy-internal likelihood contrasts rather than relabeled trajectories or new data. \method{} converts this contrast into bounded, response-normalized credit weights. These weights redistribute the original response-level GRPO advantage across tokens, so that rubric-dependent tokens receive more of the signed update signal without changing its direction. Unlike methods that train a token-level relevance model, \method{} requires no separate relevance-learning stage and does not change the response-level reward. By modifying only the within-response credit allocation while preserving response-level supervision, \method{} improves training effectiveness while preserving the simplicity of the GRPO.

\begin{figure*}[t]
\centering
\includegraphics[width=0.9\textwidth]{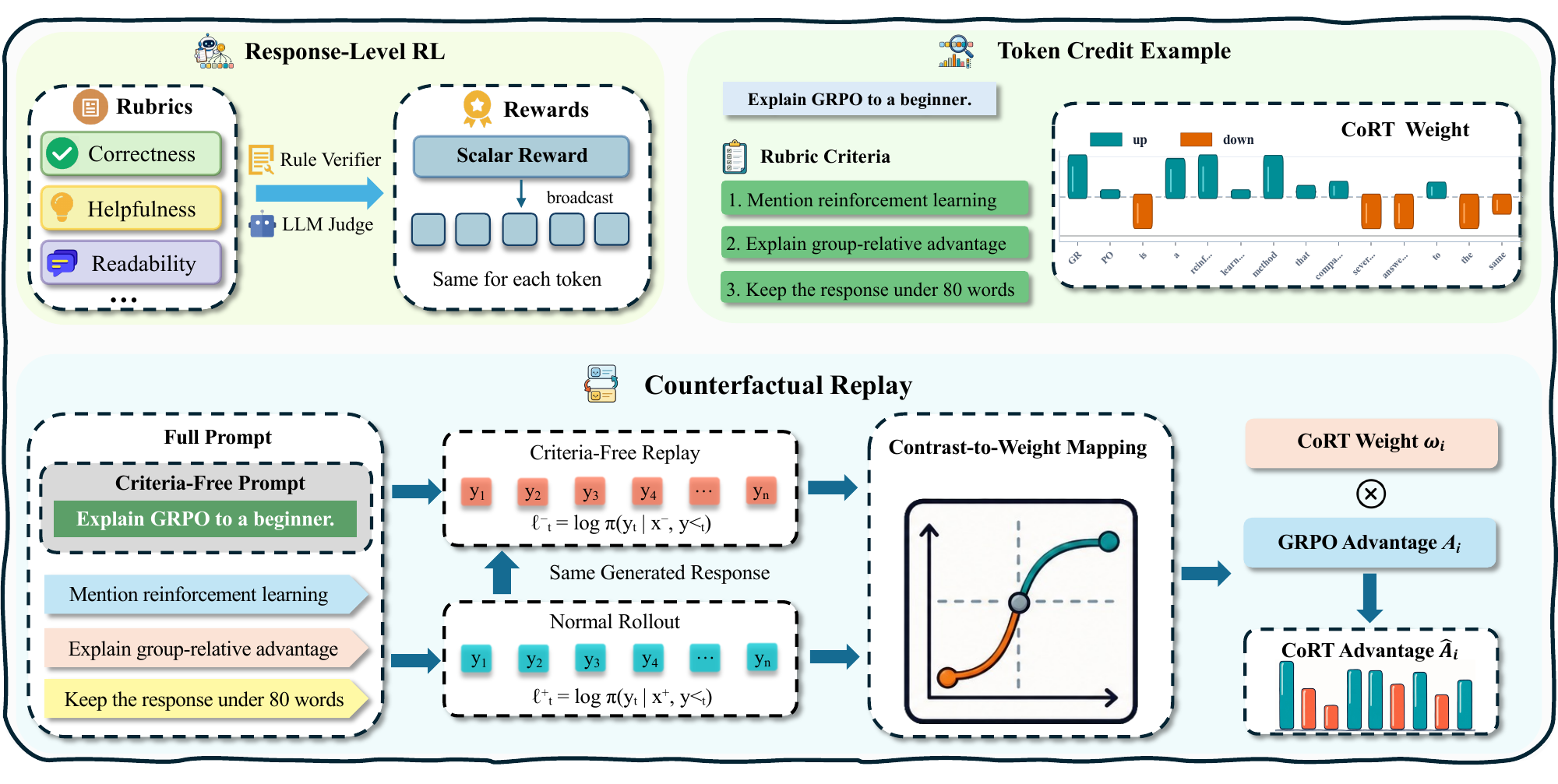}
\caption{Overview of \method{}. Response level rubric rewards are first converted into a scalar GRPO advantage, which is normally broadcast uniformly to all generated tokens. \method{} replays the same response under a criteria-free counterfactual prompt, contrasts token log probabilities against the normal rollout, maps the contrast into token weights, and uses these weights to redistribute the response level advantage across tokens.}
\label{fig:overview}
\end{figure*}

\section{Related Work}

\paragraph{Instruction following and rubric feedback.}
Instruction following has been widely studied as a central goal of language model post training, with prior work improving model alignment through curated instruction data, instruction synthesis, and preference optimization~\citep{zhou2023lima,xu2024wizardlm,lou2024muffin,rafailov2023direct}. Recent benchmarks increasingly emphasize explicit constraints and fine-grained instruction diversity~\citep{zhou2023instruction,pyatkin2026generalizing,gu2024dingo}, as well as multidimensional criteria and judge-based assessment~\citep{ye2025multi,he2025advancedif}. This evaluation paradigm has further motivated rubric and verifier based training methods, which extend RLHF beyond generic preference labels by providing more structured and inspectable supervision~\citep{peng2025verif,sun2024conifer,he2024complex,zhang2025replay}. However, in many RL formulations, such structured feedback is ultimately reduced to a response level scalar reward or advantage. As a result, the criterion level information provided by rubrics or verifiers is not directly used to determine which parts of a response contribute to satisfying or violating specific requirements. Motivated by this gap, our work studies how criterion level rubric feedback can be used for within response credit assignment, allowing structured supervision to affect token level policy updates rather than only response level scoring.

\paragraph{Policy optimization for language models.}
PPO is built around a clipped policy-gradient surrogate and remains widely used for language-model alignment~\citep{schulman2017proximal,ouyang2022training}. GRPO removes the value model by normalizing rewards within a response group, and has become a common basis for scalable RL training of LLMs~\citep{shao2024deepseekmath,guo2025deepseek}. Recent variants such as DAPO and GSPO refine large scale RL training through choices in normalization, clipping, and sampling~\citep{yu2026dapo,zheng2025group}, while other work studies group normalization, multi reward settings, and credit assignment for reasoning~\citep{parthasarathi2025grpo,liu2026gdpo}. These GRPO-style methods primarily refine how response-level feedback is normalized, sampled, and optimized, but they do not specify how rubric criteria should be attributed to tokens within a response. \method{} instead operates at the credit allocation stage, redistributing the resulting group-relative advantage across tokens while leaving the reward interface and group-relative normalization unchanged.

\paragraph{Fine grained credit assignment.}
Fine-grained supervision aims to alleviate the sparsity of outcome-only rewards by assigning training signals to smaller units of behavior. Process reward methods score intermediate reasoning steps~\citep{lightman2024let,wang2024math}, while agentic RL studies sparse long-horizon credit assignment through implicit step rewards, hindsight credit assignment, and hierarchical advantage estimation~\citep{liu2025agentic,tan2026hindsight,peng2026hiper}. In instruction following, Rubrics-to-Tokens (RTT) learns a Token-Level Relevance Discriminator to convert response-level rubric outcomes into token-level rewards~\citep{xu2026rubrics}. As the closest prior work, RTT targets the same uniform-credit problem in rubric-based RL, but training its discriminator requires a nontrivial data-generation pipeline for token-level relevance supervision. \method{} keeps the goal of fine-grained credit assignment, but replaces discriminator relevance with a policy-internal likelihood contrast. This connects \method{} to context intervention and attribution methods that test whether outputs depend on specific parts of the input context~\citep{cohen2024contextcite,sarti2024quantifying,shilpika2026probabilistic}, as well as broader counterfactual views of intervention~\citep{pearl2009causality,wachter2017counterfactual}. Rather than using such contrasts only to explain completed generations, \method{} uses them to redistribute the GRPO advantage during policy training.

\section{Rubric Signals in Group-Relative RL}

Let $x$ be an instruction and $c$ a set of rubric criteria. We denote the full rubric prompt by $x^+=(x,c)$ and the criteria-free prompt by $x^-=x$. For each prompt group, the old policy samples $G$ responses $y_i \sim \pi_{\theta_{\mathrm{old}}}(\cdot \mid x^+)$. A rubric verifier assigns each response a scalar outcome reward $r_i$, and GRPO converts it into a group-relative advantage
\[
A_i=\frac{r_i-\mu}{\sigma+\epsilon_r},
\]
where $\mu$ and $\sigma$ are the mean and standard deviation of rewards in the group~\citep{shao2024deepseekmath}.

Although the policy loss is evaluated tokenwise, standard GRPO uses the same group-relative advantage $A_i$ for every generated token in response $y_i$. For token position $t$, define the policy ratio
\[
\rho_{i,t}(\theta)=
\frac{
\pi_\theta(y_{i,t}\mid x^+,y_{i,<t})
}{
\pi_{\theta_{\mathrm{old}}}(y_{i,t}\mid x^+,y_{i,<t})
}.
\]
Following the PPO-style clipped surrogate~\citep{schulman2017proximal}, let
\[
\bar{\rho}_{i,t}(\theta)
=
\operatorname{clip}\!\left(
\rho_{i,t}(\theta),
1-\epsilon_{\mathrm{clip}},
1+\epsilon_{\mathrm{clip}}
\right).
\]
The GRPO loss can be written as
\[
\mathcal{L}_{\mathrm{GRPO}}
=
-\mathbb{E}_{i,t}
\left[
\min\!\left(
\rho_{i,t}A_i,
\bar{\rho}_{i,t}A_i
\right)
\right],
\]
where the expectation is over sampled responses and their generated tokens. This objective is agnostic to which tokens express the rubric criteria: the same response-level advantage coefficient is applied across the response.

\paragraph{Advantage redistribution.}
Our goal is not to call a new verifier or change the scalar outcome reward, but to redistribute the same response-level advantage within each sampled response. Let $T_i$ be the length of response $y_i$. We require token weights $w_{i,t}$ to satisfy
\[
\frac{1}{T_i}\sum_{t=1}^{T_i} w_{i,t}=1,
\]
so that the average shaped advantage remains
\[
\frac{1}{T_i}\sum_{t=1}^{T_i} \widehat{A}_{i,t}
=
\frac{1}{T_i}\sum_{t=1}^{T_i} w_{i,t}A_i
=
A_i.
\]
Under this constraint, token weights reallocate the signed advantage across tokens while preserving the response-level reward and group-relative normalization. \method{} realizes this by deriving $w_{i,t}$ from counterfactual replay and response normalization, replacing the uniform broadcast of $A_i$ with rubric-dependent token weights.

\section{\method}

\begin{algorithm}[t]
\caption{\method{} pipeline for one prompt group}
\label{alg:core}
\begin{algorithmic}[1]
\REQUIRE Full prompt $x^+$, criteria-free prompt $x^-$, responses $\{y_i\}_{i=1}^G$, rewards $\{r_i\}_{i=1}^G$
\STATE Compute group-relative advantages $A_i$ from rewards.
\STATE Obtain full-prompt log probabilities $\ell^+_{i,t}$ on sampled responses.
\STATE Score the same response token sequences under $x^-$ to obtain $\ell^-_{i,t}$.
\STATE Compute counterfactual contrasts $\Delta_{i,t}=\ell^+_{i,t}-\ell^-_{i,t}$.
\STATE Map contrasts to bounded replay-margin scores $s_{i,t}$.
\STATE Apply scheduled response normalization to obtain $w_{i,t}$.
\STATE Optimize the clipped policy objective with $\widehat{A}_{i,t}=\sg(w_{i,t})A_i$.
\end{algorithmic}
\end{algorithm}

\method{} converts a sparse response-level GRPO advantage into dense token-level credit for the same sampled response. Figure~\ref{fig:overview} provides an overview of the pipeline, while Algorithm~\ref{alg:core} gives the corresponding training procedure. For each prompt group, the standard GRPO path first converts scalar rubric rewards into response-level advantages. \method{} keeps the sampled responses and rewards fixed, then adds a counterfactual replay pass that scores the same response tokens under the criteria-free prompt. Comparing these scores with the factual full-prompt scores yields tokenwise log-probability contrasts, which indicate which tokens depend more strongly on the rubric context. These contrasts are mapped to bounded token weights and normalized within each response before being multiplied with the original advantage. In this way, the replay signal determines how the update is distributed across tokens, while the signed GRPO advantage still determines whether those tokens are reinforced or suppressed. Thus, the method changes only within-response credit allocation while leaving the rollout, verifier reward, and response-level update direction unchanged. The result is a credit-weighting mechanism that is tied to the rubric context but still anchored by the original group-relative reward signal.

\subsection{Counterfactual Scoring}

For each sampled response $y_i$, \method{} scores the same observed token trajectory under two contexts: the original prompt with rubric criteria, $x_i^+$, and the counterfactual prompt without these criteria, $x_i^-$. It computes
\[
\begin{aligned}
\ell^+_{i,t}
&=\log \pi_{\bar{\theta}}(y_{i,t}\mid x_i^+,y_{i,<t}),
\\
\ell^-_{i,t}
&=\log \pi_{\bar{\theta}}(y_{i,t}\mid x_i^-,y_{i,<t}).
\end{aligned}
\]
Here $\bar{\theta}$ denotes the frozen scoring policy for the current batch. In implementation, $\ell^+_{i,t}$ is available from the normal rollout log probabilities, while $\ell^-_{i,t}$ is computed by replaying the same response tokens under the counterfactual prompt $x_i^-$ and evaluating their per token log probabilities. The replay contrast $\Delta_{i,t}=\ell^+_{i,t}-\ell^-_{i,t}$ compares the log probability assigned to the same next token under the criteria-conditioned and criteria-free contexts. Since the prefix and target token are fixed, positive contrasts indicate stronger dependence on the rubric criteria relative to the criteria-free baseline, while small or negative contrasts indicate weak or reversed dependence. We use this contrast as a proxy for rubric dependence, rather than as a calibrated measure of token importance.

\subsection{Replay-Margin Token Credit}

The replay contrasts $\Delta_{i,t}$ are unbounded and can be heavy-tailed or model-scale dependent. We therefore map them to a bounded score before constructing token credit weights:
\[
\begin{aligned}
s_{i,t}
&=\operatorname{sigmoid}(\tau(\Delta_{i,t}-b))-\frac{1}{2},\\
\tilde{w}_{i,t}
&=1+\eta\lambda_k s_{i,t}.
\end{aligned}
\]
Here $b$ centers the replay contrast and $\tau$ controls the sharpness of the sigmoid transform. The resulting score $s_{i,t}$ lies in $(-\frac{1}{2},\frac{1}{2})$ and becomes zero when $\Delta_{i,t}=b$. The schedule coefficient $\lambda_k\in[0,1]$ controls how strongly replay weighting is activated at trainer step $k$, while $\eta$ controls the overall weighting strength, giving provisional weights in $(1-\frac{\eta\lambda_k}{2},1+\frac{\eta\lambda_k}{2})$.

\subsection{Scheduled Response Normalization}

\method{} does not use the replay weights at full strength from the first update. Let $k$ denote the trainer step and define the normalized progress $u_k$ after warmup as
\[
\begin{aligned}
u_k
&=
\operatorname{clip}\!\left(\frac{k-k_0}{K},0,1\right),\\
\lambda_k
&=
3u_k^2-2u_k^3.
\end{aligned}
\]
where $k_0$ is the warmup offset and $K$ is the ramp length. This is the standard cubic Hermite SmoothStep schedule~\citep{perlin1985image}, satisfying $\lambda_k=0$ before the ramp, $\lambda_k=1$ after the ramp, and zero slope at both endpoints. The schedule keeps early updates close to GRPO and smoothly introduces token-level weighting only after the rollout and reward statistics become more stable.

The resulting provisional weights are normalized within each response:
\[
\bar{w}_i=\frac{1}{T_i}\sum_{t=1}^{T_i}\tilde{w}_{i,t},
\qquad
w_{i,t}=\frac{\tilde{w}_{i,t}}{\bar{w}_i}.
\]
By construction, $\frac{1}{T_i}\sum_{t=1}^{T_i}w_{i,t}=1$ and $\frac{1}{T_i}\sum_{t=1}^{T_i}w_{i,t}A_i=A_i$. This keeps the tokenwise coefficients on the original GRPO advantage scale while redistributing the signed advantage across response tokens.

\subsection{Policy Objective}

The token-shaped advantage is
\[
\widehat{A}_{i,t}=\sg(w_{i,t})A_i,
\]
where $\sg(\cdot)$ stops gradients through the replay-derived weight. \method{} uses the same clipped surrogate form as GRPO, but replaces the uniform response-level advantage $A_i$ with the token-dependent advantage $\widehat{A}_{i,t}$:
\[
g_{i,t}(a)
=
\min\!\left(
\rho_{i,t}a,
\bar{\rho}_{i,t}a
\right).
\]
The resulting policy loss is
\[
\mathcal{L}_{\method}
=
-\E_{i,t}\!\left[
g_{i,t}(\widehat{A}_{i,t})
\right],
\]
where the expectation is over sampled responses and their generated tokens. Thus, $w_{i,t}$ controls the magnitude of the signed update at each token, while the original group-relative advantage $A_i$ determines whether the token is reinforced or suppressed during the policy update.

\paragraph{Computation.}
\method{} adds one additional forward scoring pass for each sampled response under the criteria-free prompt $x_i^-$. This pass reuses the generated response tokens and computes their per-token log probabilities, without extra generation, verifier calls, rejection sampling, or token-level relevance labels. Since the factual log probabilities under $x_i^+$ are already available from rollout, the additional computation is limited to counterfactual scoring and response normalization for advantage construction. Thus, \method{} remains a lightweight modification to GRPO compared with methods that require a separate relevance-learning stage.

\section{Experimental Setup}

\paragraph{Training data.}
Training uses HIR-16k~\citep{zhang2025replay}. Each example provides a prompt without the instruction list and the corresponding instruction list, allowing us to construct the two contexts required by \method{}. We obtain $x^+$ by appending the instruction list to the end of the prompt and use it throughout the main RL loop. The original prompt without the list is used as $x^-$ only for counterfactual scoring.

\paragraph{Reward modes.}
Following RTT~\citep{xu2026rubrics}, we evaluate two rubric reward granularities: Constraint Satisfaction Rate (CSR) and All-or-Nothing (AON). Let $M$ be the number of positive criteria for a response, and let $z_j\in\{0,1\}$ indicate whether the $j$-th criterion is satisfied. CSR assigns the fraction of satisfied criteria, while AON assigns reward one only when all positive criteria are satisfied:
\[
r_{\mathrm{CSR}}
=
\frac{1}{M}\sum_{j=1}^{M} z_j,
\qquad
r_{\mathrm{AON}}
=
\mathbf{1}\!\left[\sum_{j=1}^{M} z_j=M\right].
\]
These rewards are used for RL training only; validation uses the official metrics of each benchmark.

\paragraph{Models and optimization.}
The main matrix contains Qwen3-4B-Instruct and Qwen2.5-7B-Instruct, each trained under CSR and AON~\citep{yang2025qwen3,qwen2025qwen25technicalreport}. All runs use a prompt batch size of 64, sample 8 responses per prompt, set maximum prompt and response lengths to 2048 and 4096, and use rollout temperature 1.0, top-$p$ 0.99, and top-$k$ 100. The actor learning rate is $10^{-6}$ with warmup ratio 0.03 and weight decay 0.1. We do not use an explicit reward-KL term or an auxiliary actor-KL loss. The lower and upper clipping bounds are 0.2 and 0.27, respectively.

\paragraph{\method{} instantiation.}
Unless otherwise stated, \method{} uses a zero-centered replay-margin transform with unit temperature, $\eta=0.5$, a 100-step SmoothStep ramp, and response-mean normalization. Before response normalization, this bounds the active token multiplier in $(0.75,1.25)$. The replay-derived weights are applied to positive and negative group-relative advantages, so the method changes token credit rather than the reward definition.

\paragraph{Evaluation.}
Validation uses IFBench, IFEval, MultiDimIF, and AdvancedIF~\citep{pyatkin2026generalizing,zhou2023instruction,ye2025multi,he2025advancedif}. IFBench and IFEval report strict prompt-level and instruction-level accuracy. MultiDimIF reports accuracy over multi-dimensional instruction constraints. AdvancedIF follows its judge-based protocol with \texttt{o3-mini}. For all benchmarks, all methods are evaluated with 5 sampled responses per prompt, temperature 0.7, top-$p$ 0.8, and top-$k$ 20. Following the validation protocol of RTT~\citep{xu2026rubrics}, we report the best validation score within the first 500 trainer steps for IFBench, IFEval, and MultiDimIF under this shared protocol. To make the fixed-checkpoint behavior transparent, we also report the corresponding step-500 \method{} results in Table~\ref{tab:step500-results}. To limit the cost of judge-based evaluation, AdvancedIF uses the same 5-sample decoding protocol but is evaluated only at step 500.

\section{Results}

Table~\ref{tab:main-results} reports the main validation results, organized by model family and training reward. Within each CSR or AON group, GRPO, RTT, and \method{} are compared under the same reward setting. Shaded rows denote \method{}, and bold and underlined values mark the best and second-best results within each model block. SFT, DPO, and RTT denote supervised fine-tuning, direct preference optimization, and Rubrics-to-Tokens, respectively. CSR and AON denote Constraint Satisfaction Rate and All-or-Nothing rewards.

Overall, \method{} performs strongly and consistently across the main evaluation matrix. Under the same reward family, it improves over response-level GRPO in most comparisons across models, reward settings, and benchmarks. Compared with RTT, \method{} is competitive and often stronger, while avoiding a learned token-level relevance discriminator.

\begin{table*}[!t]
\centering
\small
\setlength{\tabcolsep}{1.5pt}
\renewcommand{\arraystretch}{1.10}
\begin{tabularx}{\textwidth}{@{}lYYYYYYY@{}}
\toprule
& \multicolumn{2}{c}{IFEval} & \multicolumn{2}{c}{IFBench} & MultiDimIF & \multicolumn{2}{c}{AdvancedIF} \\
\cmidrule(lr){2-3}\cmidrule(lr){4-5}\cmidrule(lr){6-6}\cmidrule(lr){7-8}
Method & Prompt & Instruction & Prompt & Instruction & Accuracy & Overall & Rubric \\
\midrule
\multicolumn{8}{@{}l}{\emph{Qwen3-4B-Instruct}} \\
Instruct & 83.40 & 88.13 & 30.95 & 32.83 & 56.08 & 44.31 & 79.17 \\
SFT & 83.73 & 88.85 & 29.59 & 31.64 & 56.50 & 45.07 & 79.23 \\
DPO & 82.62 & 87.77 & 29.93 & 32.84 & 55.83 & 45.71 & 79.33 \\
\addlinespace[1pt]
\multicolumn{8}{@{}l}{\itshape CSR reward} \\
\addlinespace[1pt]
\quad GRPO & 84.29 & 89.21 & 32.31 & 33.43 & 74.38 & 46.02 & 79.81 \\
\quad\hspace{0.45em}+\hspace{0.12em}RTT & 85.03 & \underline{90.17} & 34.01 & 36.12 & 76.33 & \underline{48.39} & 80.77 \\
\rowcolor{coreRow}
\quad\hspace{0.45em}+\hspace{0.12em}\method{} & \textbf{86.06}\gain{+1.77} & \textbf{90.48}\gain{+1.27} & 34.49\gain{+2.18} & 36.24\gain{+2.81} & \underline{80.48}\gain{+6.10} & \textbf{49.18}\gain{+3.16} & \textbf{81.14}\gain{+1.33} \\
\addlinespace[1pt]
\multicolumn{8}{@{}l}{\itshape AON reward} \\
\addlinespace[1pt]
\quad GRPO & 82.99 & 88.61 & 32.99 & 35.52 & 74.25 & 47.29 & 79.69 \\
\quad\hspace{0.45em}+\hspace{0.12em}RTT & \underline{85.21} & 88.73 & \underline{34.69} & \textbf{37.61} & 76.75 & 47.48 & \underline{80.79} \\
\rowcolor{coreRow}
\quad\hspace{0.45em}+\hspace{0.12em}\method{} & 84.66\gain{+1.67} & 89.50\gain{+0.89} & \textbf{35.10}\gain{+2.11} & \underline{37.37}\gain{+1.85} & \textbf{81.60}\gain{+7.35} & 48.03\gain{+0.74} & 80.10\gain{+0.41} \\
\midrule
\multicolumn{8}{@{}l}{\emph{Qwen2.5-7B-Instruct}} \\
Instruct & 69.68 & 77.93 & 25.85 & 27.16 & 51.08 & 29.48 & 71.31 \\
SFT & 71.90 & 79.38 & 28.23 & 30.15 & 52.83 & 29.71 & 69.61 \\
DPO & 71.16 & 78.90 & 27.55 & 28.96 & 51.08 & 30.82 & 70.90 \\
\addlinespace[1pt]
\multicolumn{8}{@{}l}{\itshape CSR reward} \\
\addlinespace[1pt]
\quad GRPO & 78.56 & 84.41 & 28.23 & 29.85 & 66.67 & 30.91 & 72.00 \\
\quad\hspace{0.45em}+\hspace{0.12em}RTT & 81.15 & 86.33 & \underline{32.31} & \underline{34.63} & 69.67 & 33.74 & \textbf{73.72} \\
\rowcolor{coreRow}
\quad\hspace{0.45em}+\hspace{0.12em}\method{} & 81.92\gain{+3.36} & \textbf{87.51}\gain{+3.10} & \textbf{34.63}\gain{+6.40} & \textbf{36.30}\gain{+6.45} & \underline{78.52}\gain{+11.85} & \textbf{34.18}\gain{+3.27} & 73.28\gain{+1.28} \\
\addlinespace[1pt]
\multicolumn{8}{@{}l}{\itshape AON reward} \\
\addlinespace[1pt]
\quad GRPO & 79.67 & 85.13 & 31.97 & 33.13 & 69.41 & 32.10 & 72.33 \\
\quad\hspace{0.45em}+\hspace{0.12em}RTT & \textbf{82.26} & 87.29 & \underline{32.31} & 33.43 & 71.83 & 32.51 & 72.70 \\
\rowcolor{coreRow}
\quad\hspace{0.45em}+\hspace{0.12em}\method{} & \underline{82.00}\gain{+2.33} & \underline{87.36}\gain{+2.23} & 29.66\gain{-2.31} & 31.22\gain{-1.91} & \textbf{79.63}\gain{+10.22} & \underline{33.75}\gain{+1.65} & \underline{73.57}\gain{+1.24} \\
\bottomrule
\end{tabularx}
\caption{Main validation results on instruction following benchmarks. Rows are grouped by model family and reward setting. Baseline rows, including Instruct, SFT, DPO, GRPO, and RTT, are taken from RTT~\citep{xu2026rubrics}; \method{} rows are our runs under the same evaluation protocol. Deltas are vs. matched GRPO. Shaded rows indicate \method{}; bold and underlined values mark the best and second-best results within each model block.}
\label{tab:main-results}
\end{table*}

\paragraph{Main comparison.}
The most controlled comparison is within each reward family, where GRPO and \method{} use the same rollout prompts, scalar rewards, and group-relative normalization. Across CSR and AON settings, \method{} improves over response-level GRPO in most comparisons across models and benchmarks. The gains also hold on AdvancedIF under the fixed-step judge-based protocol, where \method{} improves over matched GRPO across the reported metrics. These results show that the same response-level rubric signal can become more effective when its advantage is redistributed across tokens, without changing the verifier reward or group-relative normalization. 

\paragraph{Comparison with learned token relevance.}
RTT directly targets the response-to-token credit gap by training a token-level relevance discriminator~\citep{xu2026rubrics}. \method{} explores a different tradeoff by deriving token weights from policy-internal counterfactual likelihood contrasts rather than learning a separate relevance model. Across the main comparisons, \method{} is competitive with and often stronger than RTT under the same evaluation protocol, while avoiding a separate relevance-learning stage. This suggests that explicit token relevance training is not always necessary for token-level credit assignment in rubric-conditioned GRPO.

\paragraph{Qwen3-14B results.}
Table~\ref{tab:qwen14b-results} evaluates whether \method{} extends to a larger model scale. Under CSR, \method{} improves all reported metrics; under AON, it improves IFBench and MultiDimIF while trailing GRPO on IFEval. These results show that replay-based credit weighting remains effective at 14B scale, with metric-specific variation under sparse rewards.

\begin{table}[t]
\centering
\small
\setlength{\tabcolsep}{1.0pt}
\renewcommand{\arraystretch}{1.05}
\begin{tabularx}{\columnwidth}{@{}lYYYYY@{}}
\toprule
Method & \multicolumn{2}{c}{IFEval} & \multicolumn{2}{c}{IFBench} & MultiDimIF \\
\cmidrule(lr){2-3}\cmidrule(lr){4-5}
 & Prompt & Inst. & Prompt & Inst. & Acc. \\
\midrule
\addlinespace[1pt]
\multicolumn{6}{@{}l}{\itshape CSR reward} \\
\addlinespace[1pt]
\quad GRPO & 87.87 & 92.34 & 39.93 & 42.53 & 82.59 \\
\rowcolor{coreRow}
\quad\hspace{0.45em}+\hspace{0.12em}\method{} & \textbf{89.17} & \textbf{93.28} & \textbf{40.68} & \textbf{43.11} & \textbf{83.48} \\
\addlinespace[1pt]
\multicolumn{6}{@{}l}{\itshape AON reward} \\
\addlinespace[1pt]
\quad GRPO & \textbf{90.13} & \textbf{93.94} & 35.84 & 37.92 & 83.43 \\
\rowcolor{coreRow}
\quad\hspace{0.45em}+\hspace{0.12em}\method{} & 89.80 & 93.48 & \textbf{38.50} & \textbf{41.09} & \textbf{84.08} \\
\bottomrule
\end{tabularx}
\caption{Qwen3-14B results.}
\label{tab:qwen14b-results}
\end{table}

\paragraph{Compatibility with policy optimization objectives.}
Because \method{} redistributes token credit after the response-level advantage is computed, it can be combined with different policy optimization objectives. We integrate \method{} into DAPO~\citep{yu2026dapo} and GSPO~\citep{zheng2025group} under the CSR reward on Qwen3-4B-Instruct-2507. As shown in Table~\ref{tab:objective-compatibility}, adding \method{} improves all reported metrics over DAPO and improves four of the five metrics over GSPO, with only a small decrease on MultiDimIF. These results indicate that replay-derived token credit is complementary to changes in the underlying policy objective.

\begin{table}[t]
\centering
\small
\setlength{\tabcolsep}{1.0pt}
\renewcommand{\arraystretch}{1.05}
\begin{tabularx}{\columnwidth}{@{}lYYYYY@{}}
\toprule
Method & \multicolumn{2}{c}{IFEval} & \multicolumn{2}{c}{IFBench} & MultiDimIF \\
\cmidrule(lr){2-3}\cmidrule(lr){4-5}
 & Prompt & Inst. & Prompt & Inst. & Acc. \\
\midrule
\multicolumn{6}{@{}l}{\emph{Qwen3-4B-Instruct}} \\
\addlinespace[1pt]
\addlinespace[1pt]
\quad DAPO & 87.10 & 88.90 & 40.27 & 45.79 & 83.41 \\
\rowcolor{coreRow}
\quad\hspace{0.45em}+\hspace{0.12em}\method{} & \textbf{87.95} & \textbf{90.20} & \textbf{41.30} & \textbf{47.02} & \textbf{84.06} \\
\addlinespace[1pt]
\addlinespace[1pt]
\quad GSPO & 86.14 & 91.10 & 35.97 & 38.74 & \textbf{82.26} \\
\rowcolor{coreRow}
\quad\hspace{0.45em}+\hspace{0.12em}\method{} & \textbf{86.54} & \textbf{91.20} & \textbf{36.11} & \textbf{39.42} & 81.96 \\
\bottomrule
\end{tabularx}
\caption{\method{} with DAPO and GSPO (CSR).} 
\label{tab:objective-compatibility}
\end{table}

\section{Training Diagnostics}

\begin{figure}[!t]
\centering
\includegraphics[width=0.85\columnwidth]{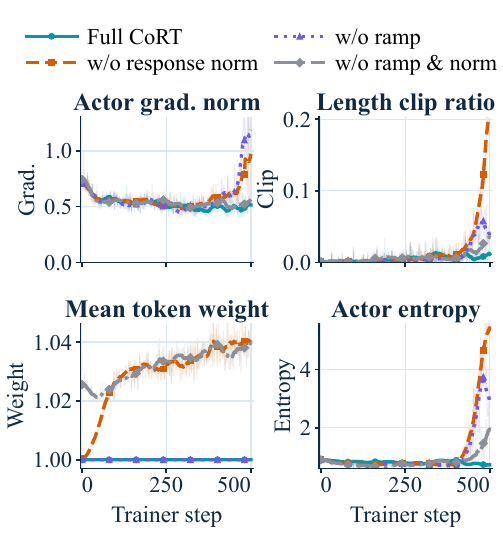}
\caption{Qwen2.5-7B CSR training diagnostics for \method{} integration variants over the first 500 trainer steps. Curves are smoothed over a 21-step window.}
\label{fig:core-stability-qwen25}
\end{figure}

Figure~\ref{fig:core-stability-qwen25} examines whether the integration components in \method{} stabilize training dynamics as intended. We use Qwen2.5-7B under the CSR reward and compare the full recipe with variants that remove response normalization, remove the SmoothStep ramp, or both. This ablation separates scale control from scheduled activation and tests whether these components keep replay-derived token weighting close to the GRPO update scale throughout the full training run.

With the full recipe, the mean token weight stays close to one, length clipping remains near zero, and the actor gradient norm and entropy stay stable. Removing response normalization causes the mean token weight to drift upward, approaching 1.04 near the end of training. This scale drift coincides with a sharp late increase in length clipping, gradient norm, and entropy, suggesting that unnormalized replay weights can act as a response-level multiplier. Removing the ramp keeps the mean token weight controlled by response normalization, but the abrupt use of replay-derived weights leads to a late gradient and entropy spike with higher length clipping than the full recipe. The variant without both controls also shows elevated mean weights and higher length clipping than the full recipe. These patterns indicate that the two controls address different stability issues: response normalization controls coefficient scale, while scheduled activation reduces abrupt changes in the token-weighted update.

\begin{figure}[t]
\centering
\includegraphics[width=0.85\columnwidth]{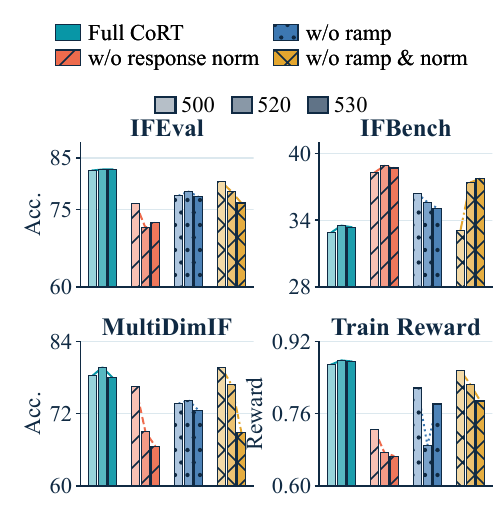}
\caption{Post-500 continuation for Qwen2.5-7B CSR integration ablations. Validation reports internal \texttt{mean@5} accuracy at steps 500, 520, and 530; training reward is a 10-step trailing mean ending at each checkpoint.}
\label{fig:qwen25-post500-validation}
\end{figure}

The post-500 continuation in Figure~\ref{fig:qwen25-post500-validation} gives the same message from the validation side. The full recipe keeps the strongest validation average and the highest 10-step trailing training reward around the nominal 500-step horizon, while the ablations show larger checkpoint-to-checkpoint movement. IFBench is the main exception: some ablations obtain higher IFBench accuracy even as their reward, IFEval, or MultiDimIF behavior is weaker. This is consistent with IFBench being sensitive to local constraint satisfaction under CSR, and suggests that single-metric gains are not enough to replace integration controls. Values are reported in Table~\ref{tab:qwen25-csr-post500-validation}.

Overall, these diagnostics support the integration design of \method{} in this setting. The full recipe is the only variant that keeps the average token coefficient on the GRPO scale while maintaining low length clipping and stable gradient norms, and it also preserves the strongest average validation and training reward around the 500-step horizon. The ablations are not strictly monotonic because the two controls address different failure modes: response normalization controls update scale, while the ramp controls how abruptly token-level redistribution is introduced. This suggests that \method{} works best when replay-derived token weights are both response-normalized and introduced gradually throughout the run.

\section{Conclusion}

We presented \method{}, a lightweight credit assignment method for rubric-based instruction following RL. The central observation is that response-level criteria rewards are optimized through tokenwise policy updates, yet GRPO assigns the same response advantage to all generated tokens. \method{} addresses this within-response mismatch by using counterfactual replay to rescore sampled responses after removing the criteria, and by converting the resulting likelihood contrast into normalized token credit weights. These weights redistribute the signed GRPO advantage toward tokens that depend more strongly on the criteria context, while keeping the original rollout distribution, verifier, scalar reward, and clipped GRPO surrogate structure intact without training a separate token relevance model. The bounded replay-margin transform, response normalization, and scheduled activation further make this redistribution stable in practice.

Across instruction following benchmarks and reward granularities, \method{} improves over response-level GRPO in most comparisons and remains competitive with learned token relevance methods. Its consistent gains when combined with DAPO and GSPO further show that replay-derived token credit is not specific to standard GRPO and can be integrated with different policy optimization algorithms. The training diagnostics further show that response normalization and scheduled activation help keep token weights on the GRPO advantage scale and reduce unstable training behavior. These results support the view that structured criteria can be useful not only for scoring responses, but also for guiding where credit is assigned within a response. Since \method{} estimates criteria reliance along observed trajectories, its signal can be weaker when criteria have little effect on token likelihoods, when they are redundant with the original prompt, or when the response-level reward is noisy. Future work can explore finer-grained interventions, criterion-specific allocation, and longer-horizon credit across turns, actions, and tool calls.

\bibliography{references}

\clearpage
\appendix

\section{Additional Results}

\subsection{Fixed Step-500 \method{} Checkpoints}

The main table selects the best official validation checkpoint within the first 500 trainer steps for IFBench, IFEval, and MultiDimIF. Table~\ref{tab:step500-results} reports the corresponding \method{} results at exactly step 500. This table is included to make the fixed-checkpoint behavior transparent; it is not used to replace the best-within-window selection rule in the main comparison.

\begin{center}
\small
\setlength{\tabcolsep}{1.8pt}
\begin{tabular}{lccccc}
\toprule
Method & \multicolumn{2}{c}{IFEval} & \multicolumn{2}{c}{IFBench} & MultiDimIF \\
\cmidrule(lr){2-3}\cmidrule(lr){4-5}
 & Prompt & Instruction & Prompt & Instruction & Accuracy \\
\midrule
\multicolumn{6}{l}{Qwen3-4B} \\
\method{}-CSR & 86.06 & 90.48 & 34.49 & 36.24 & 80.48 \\
\method{}-AON & 84.66 & 89.50 & 32.31 & 34.44 & 81.60 \\
\midrule
\multicolumn{6}{l}{Qwen2.5-7B} \\
\method{}-CSR & 81.77 & 87.36 & 34.63 & 36.30 & 78.52 \\
\method{}-AON & 81.81 & 87.29 & 29.32 & 30.27 & 78.62 \\
\bottomrule
\end{tabular}
\captionof{table}{\method{} results at exactly step 500 for the main experiment matrix.}
\label{tab:step500-results}
\end{center}

\subsection{General Capability Checks}

Table~\ref{tab:general-capability} reports mean@5 evaluation on Math500, GPQA-Diamond, and MMLU-Pro for the original instruction-tuned models and the step-500 \method{} checkpoints. These evaluations use the same original instruction-tuned checkpoint for the CSR and AON comparisons; only the \method{} checkpoint changes with the reward granularity. This provides a complementary check on whether instruction-following optimization changes the model's general math and knowledge capabilities. Across both model families, \method{} largely preserves this general capability profile.

\begin{center}
\small
\setlength{\tabcolsep}{2.0pt}
\begin{tabularx}{0.98\columnwidth}{lYYYY}
\toprule
Method & Math500 & GPQA & MMLU-Pro & Avg. \\
\midrule
\multicolumn{5}{l}{Qwen3-4B} \\
Instruct & 86.28 & 57.98 & 70.16 & 71.47 \\
\method{}-CSR & \stackgain{86.52}{+0.24} & \stackgain{59.09}{+1.11} & \stackgain{69.62}{-0.54} & \stackgain{71.74}{+0.27} \\
\method{}-AON & \stackgain{86.36}{+0.08} & \stackgain{57.78}{-0.20} & \stackgain{70.18}{+0.02} & \stackgain{71.44}{-0.03} \\
\midrule
\multicolumn{5}{l}{Qwen2.5-7B} \\
Instruct & 70.64 & 33.23 & 55.48 & 53.12 \\
\method{}-CSR & \stackgain{69.32}{-1.32} & \stackgain{33.84}{+0.61} & \stackgain{54.34}{-1.14} & \stackgain{52.50}{-0.62} \\
\method{}-AON & \stackgain{70.20}{-0.44} & \stackgain{33.64}{+0.41} & \stackgain{55.27}{-0.21} & \stackgain{53.04}{-0.08} \\
\bottomrule
\end{tabularx}
\captionof{table}{General capability checks under mean@5 evaluation. Deltas are relative to the original instruction-tuned model within each family.}
\label{tab:general-capability}
\end{center}

\subsection{Relation to Self-OPD}

On-policy self-distillation and \method{} share the goal of making sparse response-level supervision produce denser token-level training signals on trajectories sampled from the current policy. Self-Distilled RLVR derives token-level policy differences from a privileged context while keeping the RLVR reward as the update direction~\citep{yang2026self}. TRACE further routes self-OPD signals to critical spans, reducing updates on redundant tokens and limiting privileged-information leakage~\citep{wang2026trace}. \method{} differs in the source and role of the dense signal. It does not use a privileged answer trace, span annotation, or external target distribution. Instead, it keeps the sampled response fixed, removes the rubric criteria from the prompt, and treats the criteria-free log probability as a background likelihood baseline. The resulting full-vs.-criteria-free contrast is then multiplied by the signed GRPO advantage: successful responses reinforce tokens whose likelihood depends on the rubric context, while unsuccessful responses suppress such tokens more strongly. Thus, \method{} is closer to signed counterfactual contrast shaping than to self-distillation, even though both approaches use on-policy token-level modulation.

\subsection{Replay Case Studies}

We include several HIR replay examples to clarify what the replay contrast measures on actual generated responses. These examples serve three purposes. First, standard replay cases show whether high-contrast tokens correspond to visible criteria-controlled spans. Second, matched context controls test whether the signal is explained only by removing prompt text or by changing the length of the prompt. Third, criterion-wise removal checks whether removing a specific criterion produces localized changes on the tokens that realize that criterion. Together, these examples show that replay contrast is a policy-internal sensitivity signal to criteria.

The first case uses a HIR prompt asking why a balanced diet matters for athletes. The criteria include several visible formatting requirements, including English output, at least five sentences, exactly three Markdown bullet points, a postscript beginning with \texttt{P.S.}, a fixed final phrase, and enclosing double quotation marks. We sample a response from the step-500 Qwen3-4B \method{} checkpoint under the full criteria-conditioned prompt and replay the same fixed response after removing the criteria.

The text below shows the prompt, criteria, and response excerpt used in the replay analysis. We abbreviate non-critical explanatory content with ellipses, while preserving the visible criteria-controlled spans that appear in Table~\ref{tab:replay-case-study}.

\begin{quote}
\small
\textbf{Prompt.} Provide a detailed response to the following query: ``Explain the importance of maintaining a balanced diet for athletes.''

\smallskip
\noindent\textbf{Rubric criteria.} The response should be in English, contain at least five sentences, contain exactly three Markdown bullet points, add a postscript beginning with \texttt{P.S.}, finish with the exact phrase ``Is there anything else I can help with?'', and wrap the entire response with double quotation marks.

\smallskip
\noindent\textbf{Generated response excerpt.} ``Maintaining a balanced diet is crucial for athletes because it directly impacts their performance, recovery, and overall health. A well-balanced diet provides the necessary nutrients that fuel physical activity, support muscle repair, and enhance immune function. \ldots

Here are three key reasons why a balanced diet is important for athletes:

\texttt{*} This is point 1. It ensures consistent energy supply during prolonged physical activity.\\
\texttt{*} This is point 2. It supports muscle growth and repair through adequate protein intake.\\
\texttt{*} This is point 3. It strengthens the immune system and promotes better overall health.

\ldots

\texttt{P.S.} Remember that individual nutritional needs may vary based on sport, age, gender, and activity level, so athletes should consult with a sports nutritionist to tailor their diet effectively. Is there anything else I can help with?''
\end{quote}

Table~\ref{tab:replay-case-study} reports representative high-contrast tokens from the same fixed response. The selected tokens align with the visible requirements in the criteria: the opening quotation mark realizes the wrapping constraint, \texttt{three} reflects the required number of bullet points, \texttt{*} marks the Markdown bullet structure, \texttt{P} and \texttt{.S} realize the required postscript prefix, and \texttt{Is} and \texttt{?\char34} correspond to the fixed final phrase and closing quote. In contrast, these tokens are not simply generic content words about nutrition or athletes. This case therefore shows that the replay contrast highlights positions whose likelihood is tied to the removed criteria, supporting its use as a policy-internal proxy for criteria dependence.

\begin{center}
\small
\setlength{\tabcolsep}{3.0pt}
\begin{tabularx}{0.98\columnwidth}{lXcc}
\toprule
Token & Rubric cue & $\Delta_t$ & $w_t$ \\
\midrule
\texttt{\char34} & opening quotation mark & 26.37 & 1.19 \\
\texttt{three} & exact number of bullet points & 16.96 & 1.19 \\
\texttt{*} & Markdown bullet marker & 20.75 & 1.19 \\
\texttt{P} & start of postscript & 12.62 & 1.19 \\
\texttt{.S} & postscript continuation & 7.47 & 1.19 \\
\texttt{Is} & start of fixed final phrase & 13.61 & 1.19 \\
\texttt{?\char34} & final phrase and closing quote & 3.17 & 1.17 \\
\bottomrule
\end{tabularx}
\captionof{table}{Replay contrasts from the balanced-diet HIR case generated by the step-500 Qwen3-4B \method{} checkpoint.}
\label{tab:replay-case-study}
\end{center}

We next use a pandemic-economy prompt as a control-friendly case. Unlike the balanced-diet example, this prompt contains several separable local requirements, including section headers, Markdown bullets, highlighted spans, and the keyword \texttt{pandemic}. This makes it suitable for the matched-context and criterion-wise removal controls below. Table~\ref{tab:pandemic-replay-case} first verifies the basic replay pattern on this response: high-contrast tokens appear on the required output elements before we apply the control replays.

\begin{center}
\small
\setlength{\tabcolsep}{3.0pt}
\begin{tabularx}{0.98\columnwidth}{lXcc}
\toprule
Token & Rubric cue & $\Delta_t$ & $w_t$ \\
\midrule
\texttt{Section} & start of required section label & 34.88 & 1.17 \\
\texttt{\char42} & highlighted-span marker & 23.59 & 1.17 \\
\texttt{highlight} & highlighted-span content & 14.11 & 1.17 \\
\texttt{-} & Markdown bullet marker & 11.31 & 1.17 \\
\texttt{point} & bullet-list template word & 11.70 & 1.17 \\
\texttt{pandemic} & required repeated keyword & 4.35 & 1.17 \\
\bottomrule
\end{tabularx}
\captionof{table}{Replay contrasts from the pandemic HIR case generated by the step-500 Qwen3-4B \method{} checkpoint.}
\label{tab:pandemic-replay-case}
\end{center}

We next run matched context controls on the same pandemic response to test whether the signal is mainly an artifact of prompt removal. Table~\ref{tab:pandemic-context-controls} compares the criteria-free replay with two alternatives: replacing the criteria with neutral filler of similar length, and replacing them with criteria from another HIR example. In all three settings, visible criteria-controlled tokens have larger average contrast than other tokens. This suggests that the observed signal is not explained only by prompt length or by the presence of an instruction block.

\begin{center}
\small
\setlength{\tabcolsep}{2.4pt}
\begin{tabularx}{0.98\columnwidth}{lccX}
\toprule
Control context & Cue $\bar{\Delta}$ & Other $\bar{\Delta}$ & Representative high-contrast tokens \\
\midrule
Criteria-free prompt & 6.11 & 1.02 & \texttt{Section}, \texttt{*}, \texttt{highlight}, \texttt{-} \\
Length-matched neutral filler & 6.67 & 1.20 & \texttt{Section}, \texttt{*}, \texttt{highlight}, \texttt{-} \\
Shuffled HIR criteria & 5.76 & 0.97 & \texttt{Section}, \texttt{-}, repeated section markers \\
\bottomrule
\end{tabularx}
\captionof{table}{Matched context controls for the pandemic case. Cue tokens are visible criteria-controlled tokens such as section labels, Markdown markers, highlight markers, and the required keyword.}
\label{tab:pandemic-context-controls}
\end{center}

Criterion-wise removal provides a more local control. We rescore the same fixed response after removing one criterion at a time from the full criteria list. Table~\ref{tab:pandemic-removal-controls} shows that local criteria produce localized effects. Removing the bullet criterion emphasizes bullet markers and bullet template words; removing the highlight criterion emphasizes emphasis markers and \texttt{highlight}; and removing the section criterion emphasizes \texttt{Section}. In contrast, removing global criteria such as English language or minimum sentence count has little concentrated effect on visible cue tokens. These controls support the interpretation that replay contrast captures criterion-dependent likelihood changes when a criterion leaves a local trace in the response, while global or absence-style criteria may remain diffuse.

\begin{center}
\small
\setlength{\tabcolsep}{2.0pt}
\begin{tabularx}{0.98\columnwidth}{lccX}
\toprule
Removed criterion & Cue $\bar{\Delta}$ & Other $\bar{\Delta}$ & Highest affected tokens \\
\midrule
At least five sentences & 0.07 & 0.11 & newline and generic opening tokens \\
English language & -0.01 & -0.00 & no concentrated rubric cue \\
Keyword \texttt{pandemic} & 0.67 & 0.21 & \texttt{pandemic} (6.36, 4.08), nearby temporal words \\
Exactly two bullets & 2.29 & 0.37 & \texttt{-} (35.87), \texttt{point} (18.74), \texttt{This} (12.13) \\
Highlighted spans & 4.67 & 0.37 & \texttt{*} (27.76, 23.33), \texttt{highlight} (14.24) \\
Two section labels & 3.85 & 0.27 & \texttt{Section} (43.25, 18.09) \\
\bottomrule
\end{tabularx}
\captionof{table}{Criterion-wise removal controls on the pandemic case. Values are computed as full-prompt log probability minus the log probability after removing one criterion.}
\label{tab:pandemic-removal-controls}
\end{center}

\subsection{Post-500 Training and Validation Continuation}

Table~\ref{tab:qwen25-csr-post500-validation} reports the values behind Figure~\ref{fig:qwen25-post500-validation}. Validation uses the internal \texttt{mean@5} fields from the Qwen2.5-7B CSR integration-ablation runs. IFEval, IFBench, and MultiDimIF are evaluated at selected trainer steps, and the average is the unweighted mean of these three accuracies. Training reward is reported as a 10-step trailing mean ending at each selected step, rather than as a single rollout batch.

\begin{center}
\small
\setlength{\tabcolsep}{1.8pt}
\begin{tabular}{llrrrrr}
\toprule
Variant & Step & IFEval & IFBench & MultiDimIF & Avg. & Train R. \\
\midrule
\rowcolor{coreRow}
Full & 500 & 82.59 & 32.90 & 78.33 & 64.61 & 0.869 \\
\rowcolor{coreRow}
 & 520 & 82.77 & 33.52 & 79.72 & 65.34 & 0.878 \\
\rowcolor{coreRow}
 & 530 & 82.77 & 33.38 & 78.04 & 64.73 & 0.876 \\
\addlinespace[1pt]
w/o norm & 500 & 76.19 & 38.29 & 76.56 & 63.68 & 0.724 \\
 & 520 & 71.46 & 38.91 & 68.96 & 59.78 & 0.674 \\
 & 530 & 72.42 & 38.70 & 66.57 & 59.23 & 0.666 \\
\addlinespace[1pt]
w/o ramp & 500 & 77.67 & 36.38 & 73.68 & 62.58 & 0.817 \\
 & 520 & 78.48 & 35.63 & 74.21 & 62.77 & 0.690 \\
 & 530 & 77.49 & 35.09 & 72.59 & 61.72 & 0.781 \\
\addlinespace[1pt]
w/o both & 500 & 80.37 & 33.11 & 79.67 & 64.38 & 0.856 \\
 & 520 & 78.48 & 37.41 & 76.82 & 64.24 & 0.825 \\
 & 530 & 76.27 & 37.75 & 68.93 & 60.98 & 0.788 \\
\bottomrule
\end{tabular}
\captionof{table}{Qwen2.5-7B CSR post-500 continuation for \method{} integration ablations. Validation values are internal \texttt{mean@5} accuracies at the selected trainer steps; training reward is a 10-step trailing mean.}
\label{tab:qwen25-csr-post500-validation}
\end{center}

\subsection{Failure-Mode Diagnostics}

\begin{figure}[!t]
\centering
\includegraphics[width=0.95\columnwidth]{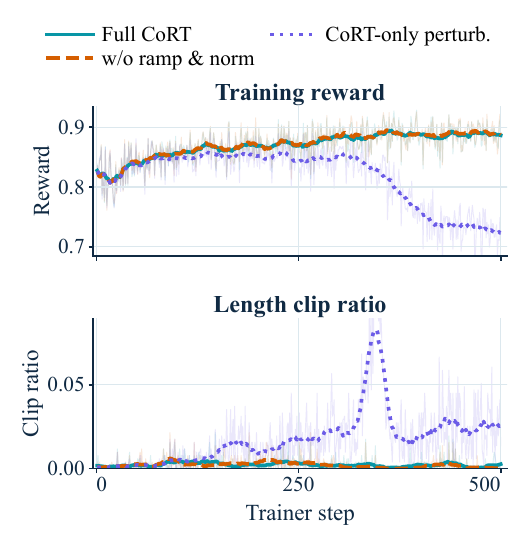}
\caption{Qwen3-4B CSR failure-mode diagnostics for a \method{}-only perturbation variant over the first 500 trainer steps. Curves are smoothed over a 21-step window.}
\label{fig:core-failure-qwen3}
\end{figure}

Figures~\ref{fig:core-stability-qwen25} and~\ref{fig:core-failure-qwen3} summarize the diagnostic runs used to check \method{}'s integration controls. In Qwen2.5-7B CSR runs, removing response normalization leads to elevated length clipping and a larger token-weight scale, while removing the activation ramp increases late-training gradient and entropy relative to the full recipe. In Qwen3-4B CSR runs, a \method{}-only perturbation variant that applies replay-derived perturbations without the standard GRPO advantage path shows lower training reward and larger length clipping; this run was stopped shortly after step 500.

These patterns clarify how the integration controls interact. Response normalization keeps the average token multiplier on the GRPO scale, so replay-derived weights redistribute credit within a response instead of changing the response-level update magnitude. The SmoothStep ramp delays the full strength of token weighting until the early rollout and reward statistics have stabilized, reducing abrupt changes in the policy-gradient multiplier. The GRPO advantage path supplies the signed reward direction: high-advantage responses reinforce rubric-sensitive tokens, while low-advantage responses suppress them. When replay perturbations are used without this reward-conditioned direction, the token contrast can still change the policy, but the update is no longer anchored to group-relative outcome improvement. The failure-mode curves are consistent with this design choice, showing that \method{} works best as a credit allocator for an existing advantage rather than as an independent token-level training signal.

\end{document}